# Title page

**Original Research**

**Title**: A Hybrid Transfer Learning Assisted Decision Support System for Accurate Prediction of Alzheimer Disease.

**Authors' Order:**

1. Mahin Khan Mahadi
   (Corresponding Author)

   Orcid: 0000-0003-3363-2365

   Department of EEE
   Islamic University of Technology, Gazipur, Bangladesh
   Email: mahinkhan@iut-dhaka.edu.

2. Abdullah Abdullah
   Department of EEE
   Islamic University of Technology, Gazipur, Bangladesh
   Email: abdullah2@iut-dhaka.edu

3. Jamal Uddin
   Department of EEE
   Islamic University of Technology, Gazipur, Bangladesh
   Email: jamaluddin@iut-dhaka.edu

4. Asif Newaz
   Department of EEE
   Islamic University of Technology, Gazipur, Bangladesh
   Email: asifnewaz@iut-dhaka.edu

# Abstract

Alzheimer's disease (AD) is the most common long-term illness in elderly people. In recent years, deep learning has become popular in the area of medical imaging and has had a lot of success there. It has become the most effective way to look at medical images. When it comes to detecting AD, the deep neural model is more accurate and effective than general machine learning. Our research contributes to the development of a more comprehensive understanding and detection of the disease by identifying four distinct classes that are predictive of AD with a high weighted accuracy of 98.91%. A unique strategy has been proposed to improve the accuracy of the imbalance dataset classification problem via the combination of ensemble averaging models and five different transfer learning models in this study. EfficientNetB0+Resnet152(effnet+res152) and InceptionV3+EfficientNetB0+Resnet50 (incep+effnet+res50) models have been fine-tuned and have reached the highest weighted accuracy for multi-class AD stage classifications.

## 1. Introduction

AD is a progressive neurodegenerative disorder that leads to cognitive decline and functional impairment by causing the loss of nerve cells and brain tissue. It is estimated that by 2050, 1 in 85 people worldwide will be affected by this disease, leading to a significant increase in its economic burden [1, 2]. Magnetic resonance imaging (MRI) has become a useful tool for studying the pathological changes associated with Alzheimer's in living individuals [3]. Machine learning (ML) methods have been applied to neuroimaging data to develop accurate and personalized diagnostic and prognostic models for Alzheimer's disease. Convolutional neural networks (CNNs) [4] have been used in recent studies to aid in the diagnosis of Alzheimer's disease.

AD is commonly associated with short-term memory difficulties, but it can also affect other cognitive functions such as expressive speech, visuospatial processing, and mental agility. While most cases of AD are not inherited dominantly, genetics play a complex role in the development of the disease. The severity of cognitive impairment in AD varies, with early symptoms possibly being a subjective decline in mental abilities without objective cognitive testing deficits [5]. Mild cognitive impairment (MCI) [6] is the earliest symptomatic stage of cognitive impairment, where one or more cognitive domains are mildly affected while functional abilities remain relatively intact. Dementia, on the other hand, is defined as a cognitive impairment that significantly impairs independence and daily functioning. AD is typically associated with gradual onset and progressive cognitive decline with prominent amnestic symptoms. Previously, AD was considered a clinicopathological entity, meaning that the clinical syndrome of amnestic dementia was assumed to be caused by AD pathology if other conditions were ruled out [7].

Early detection of Alzheimer's disease is crucial for initiating treatments to slow down its progression. The use of computing resources in healthcare departments is increasingly prevalent, with a shift towards electronic health records (EHRs) to replace traditional paper-based forms. However, processing EHR data using conventional methods such as database management software is challenging due to its unstructured nature. ML [8] techniques can be applied to EHRs to enhance the quality and productivity of healthcare centers [9].

Deep learning(DL) algorithms [10] have emerged as helpful techniques for constructing automated systems in the current day. Researchers make extensive use of DL algorithms for a variety of analytic tasks, including classification, detection, prediction, and regression. In this paper, several different pre-trained models [11] along with ensemble averaging have been implemented. Our analysis makes use of five different pre-trained models which are Resnet50 (res50) [12], Resnet101 (res101) [13], Resnet152 (res152) [14], InceptionV3 (incep) [15], and EfficientNetB0(effnet) [16]. These models have been hybridized via ensemble averaging method for having a better accuracy compared to other state-of-the-art models.

The remaining parts of the paper are organized as shown below: Section 2 covers the works that are relevant to this research as well as the impetus behind it. In Section 3, extensive information on the datasets and procedures that were used in the study is presented in a visually appealing style. In addition, the findings of the research as well as a discussion of them can be found in Section 4, which is then followed by the conclusion in Section 5.

## 2. Background Study

Hadeer et al. [17] have proposed a framework for the early detection and classification of AD using medical image classification and deep learning techniques. The framework takes into consideration different conditions and applies transfer learning techniques and multi-class medical image classification to overcome the challenges of small datasets. The authors propose two methods for medical image classification: simple CNN architectures and transfer learning using pre-trained models such as VGG19. The proposed framework achieved high accuracy in multi-class medical image classification of AD stages (AD, EMCI, LMCI, NC). The fine-tuned VGG19 model achieved the highest accuracy of 97%, followed by the 3D-M2IC model with an accuracy of 95.17%, and the 2D-M2IC model with an accuracy of 93.6%.

Vasukidevi et al. [18] has conducted research on the use of imaging technologies, specifically MRI scans, for the early detection of AD. His work highlights the importance of imaging technology in the medical industry and the various imaging modalities used for diagnosis, including CBIR(Content-based Image Retrieval ), CT(computerized tomography), and MRI. He also discusses the use of ML and DL techniques for image recognition and classification, specifically with CNN and Capsule Network (CapsNet) [19]. The CapsNet model obtained an average validation accuracy of 95.44% for the validation set and 94.3% for the test set. The KNN model yielded a validation accuracy of 69.46%. - The CNN model obtained an average validation accuracy of 65.88% on Kaggle and ADNI datasets.

Liu et al. [20] explored the use of deep learning techniques in the diagnosis of AD. The authors proposed an approach that combines 2D and 3D CNNs(convolutional neural networks) with RNNs(recurrent neural networks) to extract both intra-slice and inter-slice features from medical images. The 2D CNNs were used to extract intra-slice features, while the RNNs were used to capture inter-slice features of the 3D images. They proposed the use of stacked RNNs to model the sequential correlations between consecutive slices and extract inter-slice features. The success rate of the proposed method for the classification of AD vs. normal controls and mild cognitive impairment vs. normal controls was 91.2% and 91.4%, respectively.

Our proposed model can advance in the field of AD detection, being capable of making the following contributions:
- Carrying out a comprehensive investigation of the AD diagnosis system by deploying different ensemble averaging pre-trained models
- Implementing various ensemble transfer learning models which can demonstrate better detection results of each class among all the designated deep learning models.

The use of the suggested paradigm for AD detection in clinical settings has the potential to considerably simplify a variety of diagnostic procedures within the healthcare system.

# 3. Methodology
## 3.1 Data Description

A public domain dataset that is also accessible on Kaggle [21] was selected for this study to be used for training the models. There are a total of 6400 MRI pictures included within the dataset. All of the pictures have been scaled down to 128 by 128 pixels each. Images acquired by magnetic resonance imaging (MRI) that have been preprocessed are included in the dataset. There are a total of four different classes. Table 1 and Figure 1 provides a rundown of the many categories that can be found in the dataset. Moreover, Figure 2 represents some sample images from our training dataset.

Table 1: Dataset Sample Types for Proposed Model

| No | Image Types | Samples |
|---|---|---|
| 1 | Mild Demented | 896 |
| 2 | Moderate Demented | 64 |
| 3 | Non Demented | 3200 |
| 4 | Very Mild Demented | 2240 |

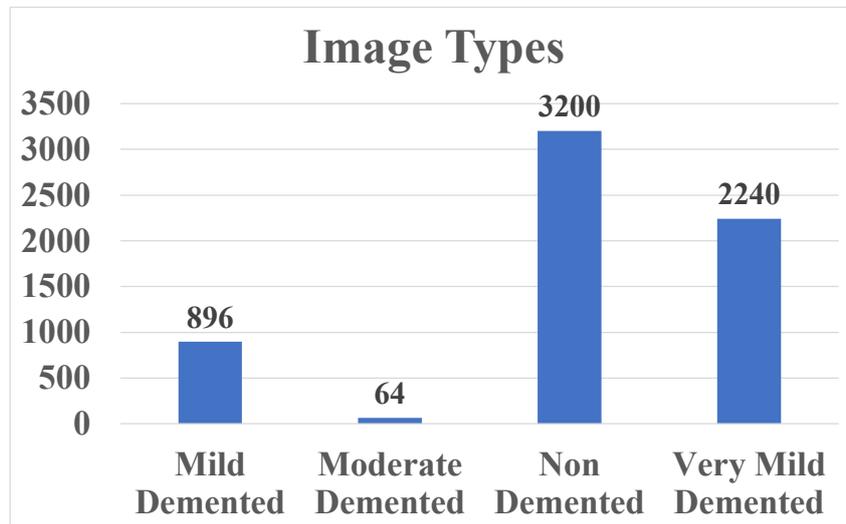

Fig 1: Dataset Image Sample Types

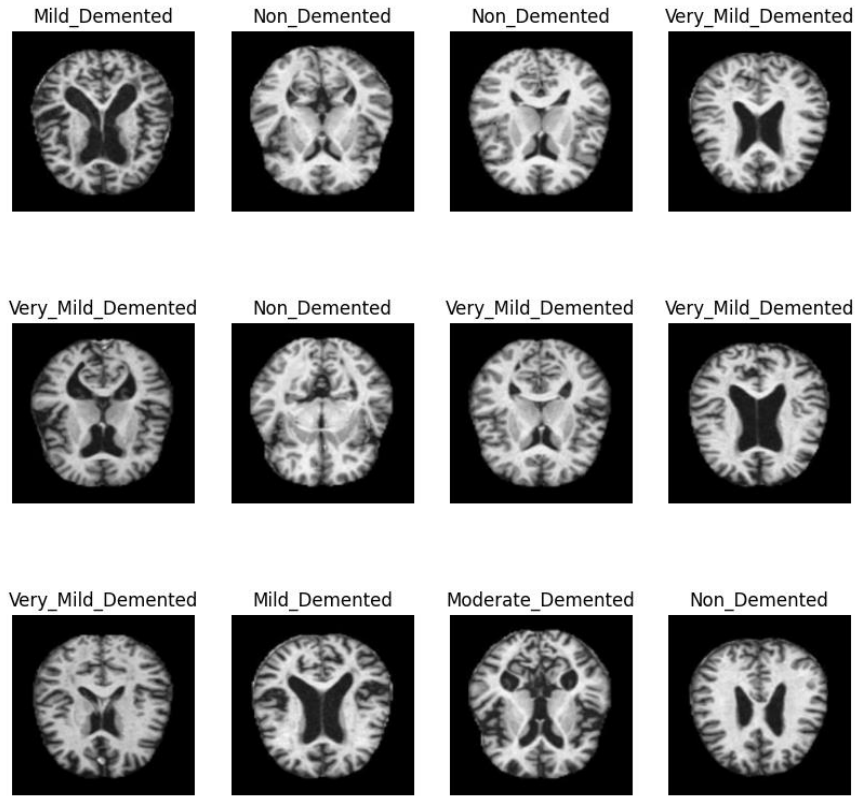

Fig 2: Sample Images from Training Data

### 3.2 Data Preprocessing

Upon acquisition of the dataset from Kaggle, it was partitioned into three distinct subsets, namely the training data, validation data, and test data, in the ratio of 80:10:10. Following the completion of prefetching on the training, validation, and test datasets, the subsequent batch of data is prepared for processing while the current batch is utilized for training purposes. This practice additionally aids in reducing the amount of idle time experienced by the central processing unit (CPU) and graphics processing unit (GPU). Subsequently, the entire dataset underwent a process of resizing and rescaling, intending to achieve normalization of pixel values across images to a uniform size and a predetermined range. In the present study, the images were rescaled through the multiplication of their height and width by a factor of 1/255. Subsequently, data augmentation is implemented to enhance the diversity and depth of the training data.

### 3.3 Training Pre-Trained Models

After preprocessing the dataset, the training data were fit into pre-trained models. A pre-trained model is a network that has already been trained, usually for an enormous image classification assignment, and then stored. If a model is trained on a sufficiently big and generic dataset, it may be used as a generic model of the visual world; this is the idea underlying transfer learning for image classification. Once these feature maps have been trained, they may be used to save time and effort over retraining a massive model on a massive dataset. In this study, several pre-trained models such as Resnet50, Resnet101, Resnet152, InceptionV3, and EfficientNetB0 have been implemented with a view to observing the performance parameters. Two additional layers (GlobalAveragePooling2D and a Dense layer) are appended onto the pre-trained trained models.

The utilization of a GlobalAveragePooling2D layer is implemented to calculate the mean value of every feature map present in the pre-trained models' output. The purpose of this particular layer is to decrease the spatial dimensions of the resulting output and generate a feature vector of fixed length. The models incorporate a Dense layer comprising of 512 units and ReLU activation. This layer performs a linear transformation on the output of the preceding layer and introduces non-linearity to the models. Furthermore, a Dropout layer is implemented with a rate of 0.3 subsequent to the Dense layer in order to mitigate overfitting. Ultimately, the models architecture incorporate a Dense layer consisting of four units and employing softmax activation as the output layer. This output layer generates a probability distribution across the four classes relevant to the given application.

### 3.3.1 Ensemble Averaging

Ensemble Averaging was applied to the pre-trained models after fitting the training data into transfer learning models. The process of averaging involves the generation of many predictions for each data point in a way that is analogous to the voting technique. During this stage of the process, an average of the predictions obtained from each of the models is calculated and then used to produce the conclusive forecast. Different tasks, including the calculation of probabilities in classification problems, may be accomplished by averaging the relevant data.

The utilization of ensemble averaging has the potential to enhance the precision of a model through the mitigation of overfitting and the augmentation of model diversity. Moreover, it has the potential to enhance the robustness of a model by mitigating the influence of outliers and data errors. As a result, it can enhance the reliability of predictions by incorporating the outputs of multiple models.

The process of ensemble averaging necessitates the training of numerous models, which may incur significant computational costs and time expenditures. It also necessitates the retention of numerous models in the computer's memory, thereby posing difficulty for models or datasets of significant size. However, in real life medical datasets tend to be limited in size, therefore it is deemed worthwhile to allocate extra resources towards the implementation of ensemble averaging, with the aim of attaining optimal performance.

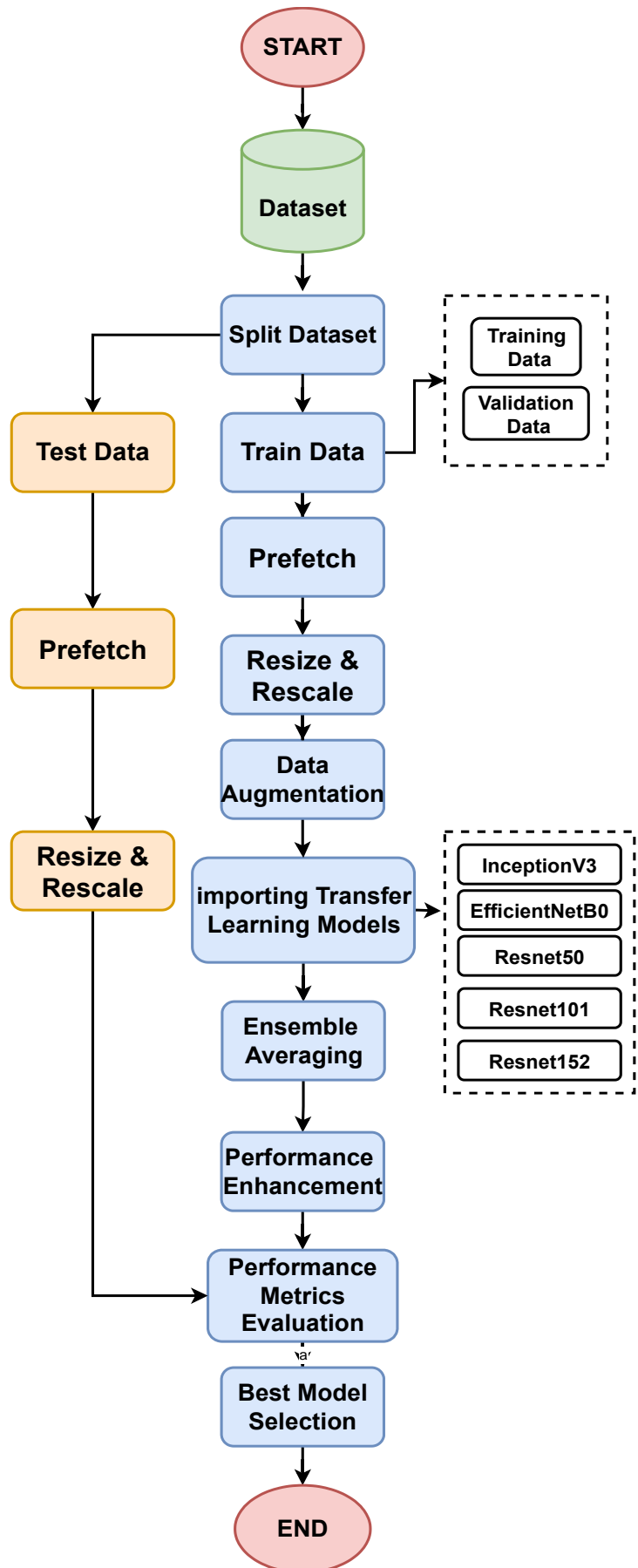

Fig 3: Overall Workflow Diagram.

# 4. Result and Analysis:
## 4.1 Performance Metrics:
Weighted accuracy is a type of performance metric used in classification problems that take into account class imbalance. In a classification problem, class imbalance occurs when the number of instances in each class is not equal. Weighted accuracy gives higher importance or weight to the accuracy of the minority class, while the majority class is given a lower weight.

$$Weighted\ Accuracy = \frac{\sum_{k=0}^{n} TP_k}{\sum_{k=0}^{n} Total\ Number\ of\ Samples_k} \quad (1)$$

Weighted precision is used to evaluate the accuracy of a classification model. It is calculated as the weighted average of the precision scores for each class in the dataset, where the weights are the number of samples in each class. Precision is the number of true positive predictions divided by the total number of positive predictions made by the model.

$$Weighted\ Precision = \frac{\sum_{k=0}^{n}(Precision_k * True\ value_k)}{\sum_{k=0}^{n} True\ value_k} \quad (2)$$

Macro precision treats each class equally regardless of its size or prevalence in the dataset, giving each class an equal weight in the calculation.

$$Macro\ Precision = \frac{\sum_{k=0}^{n} Precision_k}{n} \quad (3)$$

In micro precision, the precision is calculated by dividing the total number of true positives by the sum of true positives and false positives across all classes. It gives equal weight to each instance in the dataset, regardless of the class to which it belongs.

$$Micro\ Precision = \frac{total\ TP}{total\ TP + total\ FP} \quad (4)$$

Weighted recall is the average recall score calculated by weighting each class's recall by the number of samples in that class. In other words, it takes into account the class imbalance in the data set.

$$Weighted\ Recall = \frac{\sum_{k=0}^{n}(Recall_k * True\ value_k)}{\sum_{k=0}^{n} True\ value_k} \quad (5)$$

Macro recall is the average recall score calculated by taking the mean of recall scores for each class. It treats all classes equally, regardless of the number of samples in each class.

$$Macro\ Recall = \frac{\sum_{k=0}^{n} Recall_k}{n} \quad (6)$$

Micro recall is the overall recall score calculated by summing up the true positives for all classes and dividing it by the sum of true positives and false negatives for all classes. It treats each prediction equally, regardless of the class label.

$$Micro\ Recall = \frac{total\ TP}{total\ TP + total\ FN} \quad (7)$$

Weighted F1 takes into account the F1 score of each class and their proportion in the dataset, giving a weighted average of the F1 scores. It is a useful metric when dealing with imbalanced datasets, as it ensures that the evaluation is not biased towards the majority class.

$$Weighted\ F1 = \frac{\sum_{k=0}^{n}(F1-score_k * True\ value_k)}{\sum_{k=0}^{n} True\ value_k} \quad (8)$$

Micro F1, on the other hand, calculates the F1 score globally by aggregating the contributions of each class. It is useful when the overall performance of the classifier is of interest, and the class distribution is relatively balanced.

$$Micro\ F1 - score = 2 * \frac{Micro\ Precision * Micro\ Recall}{Micro\ Precision + Micro\ Recall} \quad (9)$$

Macro F1 calculates the F1 score for each class independently and then takes their average. It treats each class equally, regardless of its size or frequency, and is useful when the focus is on the performance of the classifier for each individual class.

$$Macro\ F1 = \frac{\sum_{k=0}^{n} F1 - score_k}{n} \qquad (10)$$

## 4.2 Performance Analysis:

Table 2: Performance metrics of ensemble averaging

| model | weighted accuracy | weighted precision | macro precision | micro precision | weighted recall | macro recall | micro recall | weighted F1 | macro F1 | micro F1 |
|---|---|---|---|---|---|---|---|---|---|---|
| Incep+Res50 | 0.9812 | 0.9819 | 0.9912 | 0.9812 | 0.9812 | 0.9481 | 0.9812 | 0.9816 | 0.9692 | 0.9812 |
| Incep+Res101 | 0.9719 | 0.9733 | 0.9870 | 0.9719 | 0.9719 | 0.9013 | 0.9719 | 0.9726 | 0.9422 | 0.9719 |
| Incep+Res152 | 0.9859 | 0.9863 | 0.9933 | 0.9859 | 0.9859 | 0.9516 | 0.9859 | 0.9861 | 0.9720 | 0.9859 |
| Effnet+Res50 | 0.9859 | 0.9862 | 0.9584 | 0.9859 | 0.9859 | 0.9498 | 0.9859 | 0.9861 | 0.9540 | 0.9859 |
| Effnet+Res101 | 0.9766 | 0.9776 | 0.9891 | 0.9766 | 0.9766 | 0.9047 | 0.9766 | 0.9771 | 0.9450 | 0.9766 |
| Effnet+Res152 | **0.9891** | 0.9892 | **0.9944** | **0.9891** | **0.9891** | 0.9542 | **0.9891** | **0.9891** | **0.9739** | **0.9891** |
| Incep+Effnet+Res50 | **0.9891** | 0.9892 | 0.9598 | **0.9891** | **0.9891** | 0.9520 | **0.9891** | **0.9891** | 0.9559 | **0.9891** |
| Incep+Effnet+Res101 | 0.9828 | 0.9858 | 0.8819 | 0.9828 | 0.9828 | 0.9421 | 0.9828 | 0.9843 | 0.9110 | 0.9828 |
| Incep+Effnet+Res152 | 0.9859 | 0.9864 | 0.9319 | 0.9859 | 0.9859 | 0.9501 | 0.9859 | 0.9862 | 0.9409 | 0.9859 |
| Res (50,101,152) | 0.9875 | **0.9895** | 0.8970 | 0.9875 | 0.9875 | 0.9491 | 0.9875 | 0.9885 | 0.9223 | 0.9875 |
| Incep+Effnet | 0.9844 | 0.9847 | 0.9922 | 0.9844 | 0.9844 | 0.9486 | 0.9844 | 0.9845 | 0.9699 | 0.9844 |
| Res (50,101) | 0.9734 | 0.9743 | 0.9467 | 0.9734 | 0.9734 | 0.9042 | 0.9734 | 0.9739 | 0.9250 | 0.9734 |
| Res (50,152) | 0.9875 | 0.9877 | 0.9591 | 0.9875 | 0.9875 | **0.9545** | 0.9875 | 0.9876 | 0.9568 | 0.9875 |
| Res (101,152) | 0.9797 | 0.9805 | 0.9905 | 0.9797 | 0.9797 | 0.9106 | 0.9797 | 0.9801 | 0.9489 | 0.9797 |
| All | 0.9828 | 0.9868 | 0.8713 | 0.9828 | 0.9828 | 0.9421 | 0.9828 | 0.9848 | 0.9053 | 0.9828 |

In terms of weighted accuracy, effnet+res152 and incep+effnet+res50 are showing the best performance which is 0.9891 whereas the performance of incep+res101 shows the lowest score compared to other models which is 0.9719.

In the case of weighted precision, the weighted precision score of res(50,101,152) is the highest among all models, which is 0.9895. The lowest score among all models is of incep+res101, which is 0.9733.

In the case of macro precision: The model effnet+res152 is showing the best performance in terms of macro precision, which is 0.9944. The lowest score among all models is of incep+effnet+res101, which is 0.8819.

In terms of micro precision, effnet+res152 and incep+effnet+res50 are showing the best performance which is 0.9891 whereas the performance of incep+res101 shows the lowest score compared to other models which is 0.9719.

Based on the weighted recall metric, it is observed that effnet+res152 and incep+effnet+res50 models show the best performance, with a score of 0.9891, whereas, the incep+res101 model shows the lowest score among all models, with a score of 0.9719.

Based on the macro recall metric, it is observed that res(50,152) model shows the best performance, with a score of 0.9545. Whereas, the incep+res101 model shows the lowest score among all models, with a score of 0.9013.

In terms of micro recall, effnet+res152 and incep+effnet+res50 are showing the best performance, which is 0.9891, whereas the performance of incep+res101 shows the lowest score compared to other models, which is 0.9719.

The results of the study show that effnet+res152 and incep+effnet+res50 are the best-performing models in terms of weighted F1 score, with a score of 0.9891. The model with the lowest weighted F1 score is incep+res101, with a score of 0.9726.

In terms of macro F1 score, effnet+res152 has the highest score of 0.9739, while incep+res101 has the lowest score of 0.9422.

In terms of micro F1, effnet+res152 and incep+effnet+res50 are showing the best performance which is 0.9891 whereas the performance of incep+res101 shows the lowest score compared to other models which is 0.9719.

The confusion matrix is a method employed to provide a concise overview of the effectiveness of a classification algorithm. The confusion matrix of ensemble averaging models is depicted in Figure 4. The images depicting the comparison between actual and predicted outcomes are presented in Figure 5. In Figure 6, the comparison of weighted accuracy of ensemble averaging models is illustrated. Table 3 and Figure 7 depict a comparison among various models that have been approached.

| Predicted Values | Mild_Demented | Moderate_Demented | Non_Demented | Very_Mild_Demented |
|---|---|---|---|---|
| Mild_Demented | 79 | 6 | 0 | 0 |
| Moderate_Demented | 0 | 6 | 1 | 0 |
| Non_Demented | 0 | 0 | 329 | 0 |
| Very_Mild_Demented | 0 | 0 | 4 | 215 |

(a) Confusion matrix of All models

| Predicted Values | Mild_Demented | Moderate_Demented | Non_Demented | Very_Mild_Demented |
|---|---|---|---|---|
| Mild_Demented | 80 | 0 | 5 | 0 |
| Moderate_Demented | 0 | 5 | 2 | 0 |
| Non_Demented | 0 | 0 | 329 | 0 |
| Very_Mild_Demented | 0 | 0 | 8 | 211 |

(b) Confusion matrix of effnet+res101

(c) Confusion matrix of effnet+res152

(d) Confusion matrix of effnet+res50

(e) Confusion matrix of incep+effnet

(f) Confusion matrix of incep+effnet+res101

(g) Confusion matrix of incep+effnet+res152

(h) Confusion matrix of incep+effnet+res50

(i) Confusion matrix of incep+res101

(j) Confusion matrix of incep+res152

(k) Confusion matrix of incep+res50

(l) Confusion matrix of res(101,152)`

(m) Confusion matrix of res(50,101,152)

(n) Confusion matrix of res(50,101)

(o) Confusion matrix of Res(50,152)

Fig 4: Confusion Matrix of Ensemble Averaging Models

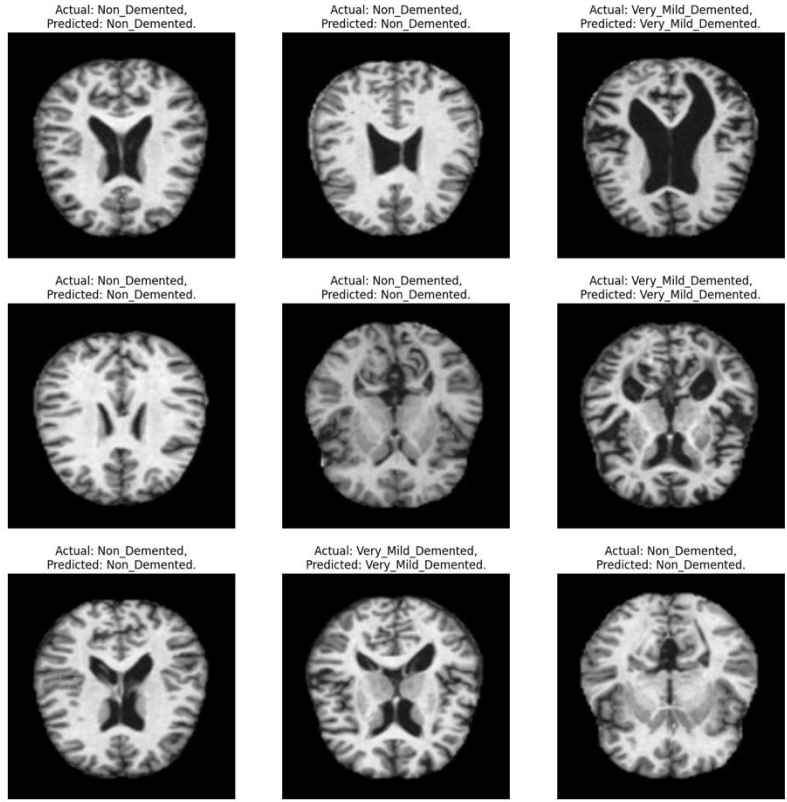

Fig 5: Actual vs Predicted Images

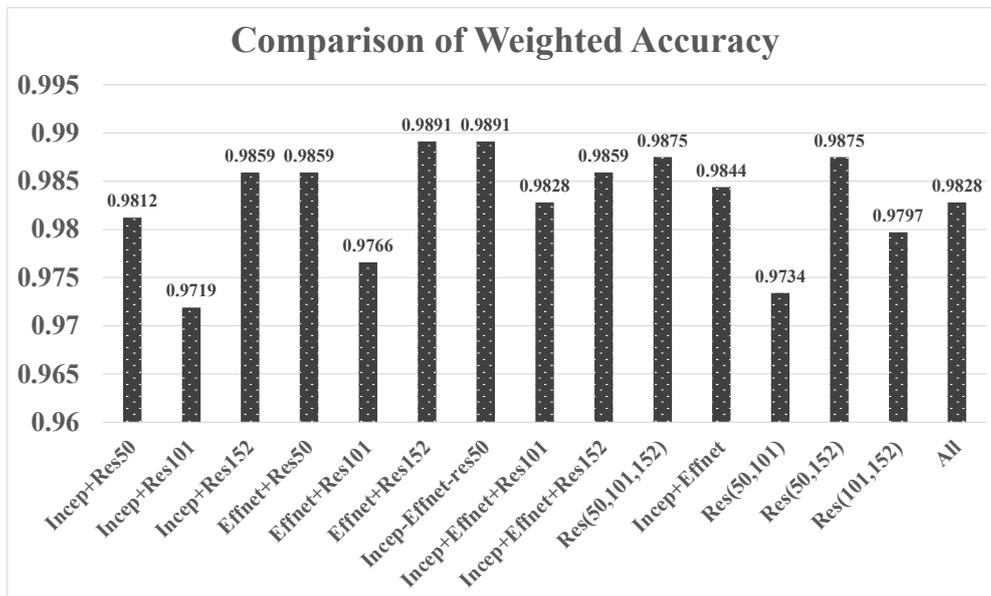

Fig 6: Comparison of Weighted Accuracy of Ensemble Averaging Models

Table 3: Comparison with other approached models

| Authors | Best Trained Models | Highest Accuracy |
|---|---|---|
| Jain et al. [22] | VGG16 | 95.73% |
| Nawaz et al. [23] | Alexnet | 92.85% |
| Sun et al. [24] | ResNet-50 | 97.10% |
| Hon et al. [25] | Inception V4 | 96.25% |
| Our Research Approach | Ensemble Averaging | 98.91% |

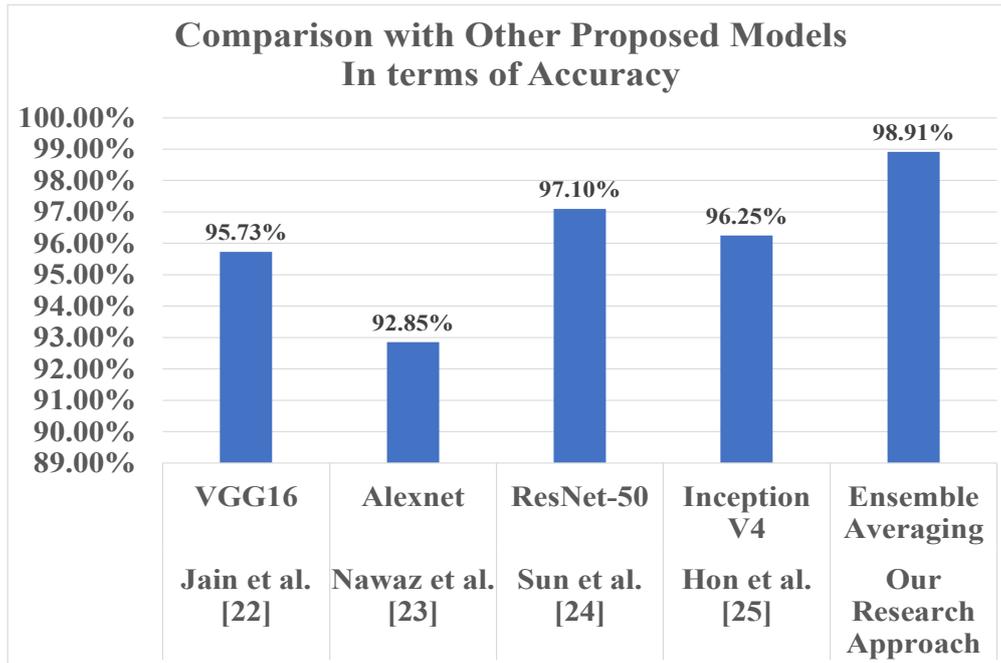

Fig 7: Comparison with Other Proposed Models

## 5. Conclusion

It has always been difficult to diagnose AD in its early stages, and associated computer researchers are continually looking for new methods to do so. Compared to other deep learning models, the performance of Ensemble Averaging Transfer Learning models is superior in classification. However, because of the lack of medical data, self-monitoring, and unsupervised approaches represent new frontiers in the study of medical images. Pre-trained models have proved successful, even though most problems associated with AD classification have not been resolved. By discovering four unique classes that predict AD with a high weighted accuracy of 98.91%, our study can assist in the creation of a more in-depth understanding of the illness. In this work, we combine ensemble averaging models with five distinct transfer learning models to provide a novel approach to boosting classification accuracy on our imbalanced dataset. In the future, we will continue our research into AD diagnostic approaches based on pre-trained models and more hybrid models.

**Data Availability**
Alzheimer MRI Preprocessed Dataset: Dataset from Kaggle was used to support this study and is available at "https://www.kaggle.com/datasets/sachinkumar413/alzheimer-mri-dataset". The dataset is cited at relevant places within the text as Ref [21].

**Competing Interests**
The authors assert that they possess no conflicting interests that could potentially impact the results of this investigation and endorse this manuscript version for dissemination.

## References

[1] Wen, Junhao, Elina Thibeau-Sutre, Mauricio Diaz-Melo, Jorge Samper-González, Alexandre Routier, Simona Bottani, Didier Dormont et al. "Convolutional neural networks for classification of Alzheimer's disease: Overview and reproducible evaluation." *Medical image analysis* 63 (2020): 101694.

[2] Rathore, Saima, Mohamad Habes, Muhammad Aksam Iftikhar, Amanda Shacklett, and Christos Davatzikos. "A review on neuroimaging-based classification studies and associated feature extraction methods for Alzheimer's disease and its prodromal stages." *NeuroImage* 155 (2017): 530-548.

[3] Qiu, Shangran, Gary H. Chang, Marcello Panagia, Deepa M. Gopal, Rhoda Au, and Vijaya B. Kolachalama. "Fusion of deep learning models of MRI scans, Mini–Mental State Examination, and logical memory test enhances diagnosis of mild cognitive impairment." *Alzheimer's & Dementia: Diagnosis, Assessment & Disease Monitoring* 10 (2018): 737-749.

[4] Albawi, Saad, Tareq Abed Mohammed, and Saad Al-Zawi. "Understanding of a convolutional neural network." In *2017 international conference on engineering and technology (ICET)*, pp. 1-6. Ieee, 2017.

[5] Knopman, David S., Helene Amieva, Ronald C. Petersen, Gäel Chételat, David M. Holtzman, Bradley T. Hyman, Ralph A. Nixon, and David T. Jones. "Alzheimer disease." *Nature reviews Disease primers* 7, no. 1 (2021): 33.

[6] Sanford, Angela M. "Mild cognitive impairment." *Clinics in geriatric medicine* 33, no. 3 (2017): 325-337

[7] Jessen, Frank, Rebecca E. Amariglio, Martin Van Boxtel, Monique Breteler, Mathieu Ceccaldi, Gaël Chételat, Bruno Dubois et al. "A conceptual framework for research on subjective cognitive decline in preclinical Alzheimer's disease." *Alzheimer's & dementia* 10, no. 6 (2014): 844-852.

[8] Hasan, Tasnimul, Md Bin Karim, Mahin Khan Mahadi, Mirza Muntasir Nishat, and Fahim Faisal. "Employment of Ensemble Machine Learning Methods for Human Activity Recognition." *Journal of Healthcare Engineering* 2022 (2022)

[9] Montagne, Christophe, Andreas Kodewitz, Vincent Vigneron, Virgile Giraud, and Sylvie Lelandais. "3D Local Binary Pattern for PET image classification by SVM, Application to early Alzheimer disease diagnosis." In *6th International Conference on Bio-Inspired Systems and Signal Processing (BIOSIGNALS 2013)*, pp. 145-150. 2013.

[10] Affonso, Carlos, André Luis Debiaso Rossi, Fábio Henrique Antunes Vieira, and André Carlos Ponce de Leon Ferreira. "Deep learning for biological image classification." *Expert systems with applications* 85 (2017): 114-122.

[11] Kabakus, Abdullah Talha, and Pakize Erdogmus. "An experimental comparison of the widely used pre-trained deep neural networks for image classification tasks towards revealing the promise of transfer-learning." *Concurrency and Computation: Practice and Experience* 34, no. 24 (2022): e7216.

[12] Mukti, Ishrat Zahan, and Dipayan Biswas. "Transfer learning based plant diseases detection using ResNet50." In *2019 4th International conference on electrical information and communication technology (EICT)*, pp. 1-6. IEEE, 2019.

[13] Lin, Shih-Lin. "Application combining VMD and ResNet101 in intelligent diagnosis of motor faults." *Sensors* 21, no. 18 (2021): 6065.

[14] Xu, Xiaoling, Wensheng Li, and Qingling Duan. "Transfer learning and SE-ResNet152 networks-based for small-scale unbalanced fish species identification." *Computers and Electronics in Agriculture* 180 (2021): 105878.

[15] Xia, Xiaoling, Cui Xu, and Bing Nan. "Inception-v3 for flower classification." In *2017 2nd international conference on image, vision and computing (ICIVC)*, pp. 783-787. IEEE, 2017.

[16] Montalbo, Francis Jesmar P., and Alvin S. Alon. "Empirical analysis of a fine-tuned deep convolutional model in classifying and detecting malaria parasites from blood smears." *KSII Transactions on Internet and Information Systems (TIIS)* 15, no. 1 (2021): 147-165

[17] Helaly, Hadeer A., Mahmoud Badawy, and Amira Y. Haikal. "Deep learning approach for early detection of Alzheimer's disease." *Cognitive computation* (2021): 1-17.

[18] Vasukidevi, G., S. Ushasukhanya, and P. Mahalakshmi. "Efficient image classification for Alzheimer's disease prediction using capsule network." *Annals of the Romanian Society for Cell Biology* (2021): 806-815.

[19] Xi, Edgar, Selina Bing, and Yang Jin. "Capsule network performance on complex data." *arXiv preprint arXiv:1712.03480* (2017).

[20] Liu, Manhua, Danni Cheng, Weiwu Yan, and Alzheimer's Disease Neuroimaging Initiative. "Classification of Alzheimer's disease by combination of convolutional and recurrent neural networks using FDG-PET images." *Frontiers in neuroinformatics* 12 (2018): 35.

[21] Alzheimer mri preprocessed dataset, Kaggle. https://www.kaggle.com/datasets/sachinkumar413/alzheimer-mri-dataset. (Accessed on 04/10/2023).

[22] Jain, Rachna, Nikita Jain, Akshay Aggarwal, and D. Jude Hemanth. "Convolutional neural network based Alzheimer's disease classification from magnetic resonance brain images." *Cognitive Systems Research* 57 (2019): 147-159.

[23] Nawaz, Hina, Muazzam Maqsood, Sitara Afzal, Farhan Aadil, Irfan Mehmood, and Seungmin Rho. "A deep feature-based real-time system for Alzheimer disease stage detection." *Multimedia Tools and Applications* 80 (2021): 35789-35807.

[24] Sun, Haijing, Anna Wang, Wenhui Wang, and Chen Liu. "An improved deep residual network prediction model for the early diagnosis of Alzheimer's disease." *Sensors* 21, no. 12 (2021): 4182.

[25] Hon, Marcia, and Naimul Mefraz Khan. "Towards Alzheimer's disease classification through transfer learning." In *2017 IEEE International conference on bioinformatics and biomedicine (BIBM)*, pp. 1166-1169. IEEE, 2017.